%% file: main.tex
\documentclass{article}
\usepackage[preprint]{neurips_2025}
\usepackage{float} 
\usepackage{amsmath} 
\usepackage{multirow} 
\usepackage{graphicx}
\usepackage[utf8]{inputenc} 
\usepackage[T1]{fontenc}    
\usepackage{hyperref}       
\usepackage{cleveref} 
\usepackage{url}            
\usepackage{booktabs}       
\usepackage{amsfonts}       
\usepackage{nicefrac}       
\usepackage{microtype}      
\usepackage{xcolor}         
\newcommand{\slora}{\textsc{SingLoRA }}

\title{SingLoRA: Low Rank Adaptation Using a Single Matrix}
\author{
  David Bensa\"id\\
  Technion - IIT \\
  Haifa, Israel\\
  \texttt{bensaiddavid@gmail.com}
  \And
Noam Rotstein\\
  Technion - IIT\\
  Haifa, Israel\\
  \And
  Roy Velich\\
  Technion - IIT\\
  Haifa, Israel
  \AND
  Daniel Bensa\"id\\
  University Paris Dauphine\\
  Paris, France
\And
  Ron Kimmel\\
  Technion - IIT\\
  Haifa, Israel
}

\begin{document}
\maketitle

\begin{abstract}
Low-Rank Adaptation (LoRA) has significantly advanced parameter-efficient fine-tuning of large pretrained models.
LoRA augments the pre-trained weights of a model by adding the product of two smaller matrices that together form a low-rank matrix update.
Recent research has shown that scale disparities between these two matrices often cause unstable training dynamics, leading to suboptimal performance.
In this paper, we propose \textbf{\slora}, which reformulates low-rank adaptation by learning the weights update as a decomposition of a \textbf{single} low-rank matrix multiplied by its transpose.
This simple design inherently removes inter-matrix scale conflicts, ensuring stable optimization, and roughly halves the parameter count. 
We analyze \slora within the infinite-width neural network framework, showing that it guarantees stable feature learning by construction. 
Extensive experiments on multiple tasks validate these benefits. 
In common sense reasoning, fine-tuning LLama 7B on MNLI with \slora achieves 91.3\% accuracy — surpassing LoRA (89.1\%) and LoRA+ (90.2\%) — while using only 60\% of their parameter budget. In image generation, fine-tuning Stable Diffusion with \slora significantly improves image fidelity on DreamBooth, achieving a DINO similarity score of 0.151, compared to scores of 0.148 and 0.143 for DoRA and LoRA, respectively.
\end{abstract}
    
\section{Introduction}

Adapting large pretrained models for specialized tasks has emerged as a central focus in machine learning research. These efforts aim to leverage the strong generalization capabilities of such models while meeting domain-specific requirements. 
Rapid scaling of model sizes and datasets has made full fine-tuning computationally prohibitive, driving the development of parameter-efficient fine-tuning (PEFT) methods to reduce computational costs. Among PEFT approaches, Low-Rank Adaptation (LoRA) \cite{hu2021lora} has gained particular popularity due to its simplicity and effectiveness in various domains. 
LoRA augments pretrained weights matrices $\mathbf{W}_0 \in \mathbb{R}^{d \times k}$ with low-rank updates, 
    \begin{equation}
    \mathbf{W}_0 + \mathbf{B}\mathbf{A}, \;
     \text{where} \; \mathbf{B} \in \mathbb{R}^{d \times r},  \; \mathbf{A} \in \mathbb{R}^{r \times k} \; \text{and} \; r \ll \min(d, k).
     \end{equation}

Recent research \cite{hayou2024loraplus, yen2024lora, zhang2024riemannian} has identified scale disparities between the matrices $\mathbf{A}$ and $\mathbf{B}$ as a fundamental challenge in LoRA. 
Such scale imbalances lead to unstable training dynamics, causing vanishing or exploding gradients and ultimately resulting in suboptimal performance. To address this challenge, we introduce \slora, which reformulates the low rank adaptation paradigm using a \textit{single} low-rank matrix $A \in \mathbb{R}^{n \times r}$ yielding the symmetric update,
\begin{equation}
\mathbf{W}_0 + \mathbf{A}\mathbf{A^\top},
\end{equation} 
This formulation provides two key benefits over traditional LoRA. First, it ensures stable optimization by design, eliminating inter-matrix scale conflicts. Second, it achieves a parameter reduction by roughly halving the number of learned parameters. Our empirical results show that these architectural improvements enable \slora to consistently exceed LoRA's performance while using significantly fewer parameters.
    
We analyze \slora in the infinite-width neural network setting \cite{hayou2019impact, hayou2024loraplus}, showing that, unlike LoRA, it guarantees stable feature learning by construction.
Building on this theoretical foundation, we extend the approach to non-square matrices, ensuring its applicability to any neural network, and validate its results through comprehensive experiments across multiple modalities.
For example, in comprehension reasoning, fine-tuning LLaMA \cite{touvron2023llama} on MNLI \cite{wang2018glue} with \slora outperforms LoRA and LoRA+ (91.3\% against  89.1\% and 90.2\% respectively) while using only 60\% of their parameter budget. In image generation, \slora increases the image fidelity of Stable Diffusion V1.5 finetuned on Dreambooth \cite{ruiz2023dreambooth} by 5.4\% compared to LoRA \cite{liu2024dora}.

\section{Related Efforts}
\subsection{LoRA}
As large multimodal models continue to scale, an increasing number of parameter-efficient fine-tuning (PEFT) techniques have been developed to facilitate their adaptation to downstream tasks. 
LoRA has emerged as one of the most popular adapters, offering efficient model adaptation across various domains. It has been studied extensively, resulting in numerous variations. DyLoRA \cite{valipour2022dylora} and AdaLoRA \cite{zhang2023adalora} focus on rank adjustment during training, with adaptive strategies to optimize rank allocation.
LoHA \cite{hyeon2021fedpara} and LoKR \cite{edalati2022krona}, propose architectural extensions and, respectively, leverage Hadamard and Kronecker products of two rank $r$ approximations to obtain more expressive adapters.
Delta-LoRA \cite{zi2023delta} modifies LoRA by updating pre-trained weights using differences in successive low-rank updates, and LoRA-Drop \cite{zhou2024lora} selects the most impactful LoRA adapters to reduce computational cost.
More recently, Weight-Decomposed Low-Rank Adaptation (DoRA) \cite{liu2024dora} decomposes the pretrained weight matrix into magnitude and direction, before employing LoRA to update the direction.
The proposed approach is complementary to these architectural extensions and can be seamlessly integrated with them, further enhancing their effectiveness.

\subsection{Stable extensions of LoRA}
A recent line of research has identified fundamental limitations in standard optimizers, such as Stochastic Gradient Descent (SGD) and Adam, to finetune LoRA modules.
LoRA+ \cite{hayou2024loraplus} shows that the matrices $A$ and $B$ should be optimized with different learning rates to ensure stable learning dynamics. Building on this theoretical foundation, Zhang \textit{et al.} \cite{zhang2024riemannian} have proposed to use Riemannian gradient descent and conditioning methods to stabilize the optimization process. Recently, LoRADoneRite \cite{yen2024lora} identified the multiplicity of possible optimizer updates for a single low-rank adapter as the source of instability in LoRA's training.
Building on these theoretical insights into optimization stability, we analyze the convergence properties of \slora. We demonstrate that its streamlined low-rank adaptation paradigm naturally promotes stable and robust optimization, without requiring careful learning rate tuning or modifications to classical optimizers.

\section{LoRA's Stability Issue}
\label{sec:lora_stability}
For a pretrained weight matrix $W_0 \in \mathbb{R}^{d \times k}$, LoRA introduces a low-rank update,
\begin{equation}
W = W_0 + \frac{\alpha}{r}BA
\end{equation}
where $B \in \mathbb{R}^{d \times r}$, $A \in \mathbb{R}^{r \times k}$ are trainable matrices with rank $r \ll \min(d,k)$, and $\alpha \in \mathbb{R}$ is a scaling factor. During fine-tuning, only $A$ and $B$ are trained while $W_0$ remains frozen. 
To preserve the pretrained model's behavior at the start of training, $B$ is initialized to zero while $A$ uses a random Gaussian initialization.
    
A recent line of work \cite{yen2024lora, hayou2024loraplus, zhang2024riemannian} has highlighted that LoRA's training dynamics often suffers from instability issues, particularly as the model width, denoted $n$, increases. 
Based on the analysis of LoRA+ \cite{hayou2024loraplus}, we examine these stability challenges in an infinite-width environment \cite{schoenholz2016deep, yang2022tensor, yang2023tensor, hayou2019impact}. 
Specifically, we investigate how the learning rate should scale so that the changes in the model output between iterations, denoted $\Delta f$, remain both bounded and non-vanishing as the network width grows. 
We refer to this property as \textit{stable features learning}, expressed mathematically as $\Delta f = \Theta_n(1)$.

\subsection{Analysis of a toy model}
In this subsection we extend the analysis of the toy model proposed in \cite{hayou2024loraplus}. For a more complete discussion of the theoretical foundations of the stable feature learning theory, we refer the reader to \cite{hayou2024loraplus, hayou2019impact}. 
Consider a linear model $f: \mathbb{R}^n \to \mathbb{R}^n$ with a rank-1 LoRA update where we have a pretrained weight matrix $W_0 \in \mathbb{R}^{n \times n}$ and trainable LoRA vectors $b, a \in \mathbb{R}^n$. The model, defined as $f(x) = (W_0 + ba^\top)x$, is trained on input-output pairs $(x,y)$,  $x, y \in \mathbb{R}^n$ with loss $
\mathcal{L} = \frac{1}{2}\|f(x) - y\|^2$.  $\mathcal{L}$ is minimized using a gradient descent method with learning rate $\eta$. 
Without loss of generality, we assume $W_0 = 0$ by defining $\tilde{y} = y - W_0x$.
It holds,
\begin{align}
    \nabla_a \mathcal{L} = (f_{t-1}(x) - y)(x^\top b_{t-1}) \quad \text{and} \quad \nabla_b \mathcal{L} = (f_{t-1}(x) - y)(a_{t-1}^\top x),
\end{align}
and at iteration $t$, the gradient updates are thus,
\begin{align}
a_t = a_{t-1} - \eta(f_{t-1}(x) - y)(x^\top b_{t-1}), \quad \text{and} \quad b_t = b_{t-1} - \eta(f_{t-1}(x) - y)(a_{t-1}^\top x).
\end{align}
To analyze stability, we examine how the network output changes between iterations,
\begin{align*}
    \Delta f_t &= f_t(x) - f_{t-1}(x)
      =  (b_t a_t^\top - b_{t-1} a_{t-1}^\top)\, x \cr &= -\underbrace{\eta\|b_{t-1}\|^2\|x\|^2[f_{t-1}(x) - y]}_{\delta_t^1}
    -  \underbrace{
    \eta(a_{t-1}^\top x)^2[f_{t-1}(x) - y]
    }_{\delta_t^2} + \underbrace{
    \eta^2\|f_{t-1}(x) - y\|^2(a_{t-1}^\top x)(b_{t-1}^\top x)x
    }_{\delta_t^3}
\end{align*}

Having $\Delta f_t = \Theta(1)$ implies that at least one of the terms $(\delta_t^i)_{i\in\{1,2,3\}}$ is $\Theta(1)$.
To ensure that updates to $a$ and $b$ significantly impact $f_t(x)$, both $\delta_t^1$ and $\delta_t^2$ must be $\Theta(1)$. 
This is because $\delta_t^1 = o(1)$ implies that the model keeps $a$ fixed, effectively training only $b$, and similarly if only $\delta_t^2 = o(1)$, only $a$ is effectively trained.
When both $\delta_t^1$ and $\delta_t^2$ are $\Theta(1)$, it follows that $\delta_t^3$ is guaranteed to be $\Theta(1)$, as shown in \cite{hayou2024loraplus}.

\subsection{Efficiency Analysis}
We analyze the scaling behavior with width using the $\gamma-$notation.
For a parameter $v$, we define,
\begin{equation}
v = \Theta(n^{\gamma[v]}),
\label{eq: gamma definition}
\end{equation}
where $\gamma[v]$ represents how $v$ scales with $n$. When applied to vectors, $\gamma[v]$ indicates that all the entries of $v$ respect Eq. \ref{eq: gamma definition}. 
For the product of vectors $v,w \in \mathbb{R}^n$, we follow a key scaling rule: $\gamma[v^\top w] = \gamma[v] + \gamma[w] + 1$, where the extra $+1$ comes from summing $n$ terms. 

As theorized in \cite{yang2022tensor, yang2023tensor}, the initialization scheme of the weights and the learning rate should be adapted as a function of the width of the network, $n$, to ensure efficient learning. 
We therefore assume that the model is trained with a gradient descent procedure with a learning rate that respects \(\eta = \Theta(n^c)\) for some \(c \in \mathbb{R}\).  \(a\) is initialized with a random Gaussian distribution scaled as \(\Theta(n^{-1/2})\) (known as Kaiming initialization \cite{he2016deep}) and $b$ is initialized as 0. Although each component of \(a_0\) is \(\Theta(n^{-1/2})\), by the Central Limit Theorem (CLT), the term \(a_0^\top x = \Theta(1)\), since it represents the sum of \(n\) independent terms with variance \(\Theta(n^{-1})\). 

Starting from initialization where $f_0(x) = 0$, efficient LoRA fine-tuning requires $\delta^1_t = \Theta(1)$, $\delta^2_t = \Theta(1)$ and $\delta^3_t = \Theta(1)$ for all $t > 1$, with $f_t(x) = \Theta(1)$ for $t > 1$. Using the $\gamma$ notation, this translates to the system,
\begin{align}
        c + 2\gamma[b_{t-1}] + 2 &= 0 \quad\text{($\delta^1_t = \Theta(1)$)} \cr
        c + 2\gamma[a_{t-1}^\top x] &= 0 \quad\text{($\delta^2_t = \Theta(1)$)} \cr
        \gamma[b_{t-1}] + \gamma[a_{t-1}^\top x] &= 0 \quad\text{($f_{t-1}(x) = \Theta(1)$)}
\end{align}
Simple calculations with this system yield $c = -1$. 
Let us analyze the updates at each step $t$. At initialization, $\gamma[a_0^\top x] = 0$ by the CLT. Using an inductive argument, for $t > 0$, the update equation for $b$ is
$b_t = b_{t-1} - \eta(f_{t-1}(x) - y)(a_{t-1}^\top x)$. Analyzing each term yields,
\begin{itemize}
    \item $(f_{t-1}(x) - y) = \Theta(1)$, as we assume that the error of the model is bounded.
    \item $a_{t-1}^\top x = \Theta(1)$ by CLT.
    \end{itemize}
Multiplying these factors, the overall update to $b$ is thus of order of $\eta$, i.e. $\Theta(n^{-1})$, implying that for $t>0$, \textbf{$b_t = \Theta(n^{-1})$}. 

Similarly for $a$, $a_t = a_{t-1} - \eta(f_{t-1}(x) - y)(x^\top b_{t-1})$. Analyzing each term yields,
\begin{itemize}
    \item $(f_{t-1}(x) - y) = \Theta(1)$.
    \item $x^\top b_{t-1} = \Theta(1)$ by the vector product rule, $\gamma[x^\top b_{t-1}] = \gamma[x] + \gamma[b_{t-1}] + 1 = 0$.
        \end{itemize}
        
Multiplying these factors, the overall update to $a$ is of order $\Theta(n^{-1})$, thus maintaining $\gamma[a_t] = -1/2$. Applying the CLT, we finally obtain \textbf{$a_t^\top x = \Theta(1)$}.

However, plugging $b_t = \Theta(n^{-1})$ and $a_t^\top x = \Theta(1)$ in the definition of $f$ yields $f_t(x) = \Theta(n^{-1})$, in contradiction with the assumption that $f_t(x) = \Theta(1)$ for efficient learning. Therefore, LoRA is not stable as updates inherently scale differently with width. This inefficiency motivates the need for alternative approaches, such as using different learning rates for $a$ and $b$, as in LoRA+ \cite{hayou2024loraplus}, or reformulating the low-rank update, as in our research. 

\section{\slora: Low-Rank Adaptation with a Single Matrix}
In this section, we first present \slora's core formulation (\ref{sec:SingLoRA_def}), showing how it achieves parameter-efficient adaptation through a symmetric low-rank update. We then analyze its training dynamics through a simplified model (\ref{sec:SingLoRA_stability}), proving that, unlike LoRA, our approach guarantees stable feature learning by construction. We establish formal convergence guarantees under standard optimizers like SGD and AdamW (\ref{sec:SingLoRA_transformation_invariance}), demonstrating that \slora eliminates the need for specialized optimization techniques. 
Finally, we extend \slora to non-square weight matrices (\ref{sec:SingLoRA_non_square}), making it applicable across diverse neural architectures. 

\subsection{\slora's formulation}
\label{sec:SingLoRA_def}
        
\slora reformulates low-rank adaptation by replacing the traditional two-matrix decomposition with a single learnable matrix. 
Given a pretrained model with frozen weight matrix $W_0 \in \mathbb{R}^{n \times n}$, \slora computes the adapted weights as,
\begin{equation}
    W_0 + \frac{\alpha}{r} u(t) AA^\top,
            \label{eq:SingLoRA_update}
        \end{equation}
where $A \in \mathbb{R}^{n \times r}$ is a low-rank trainable  matrix with rank $r \ll n$, $\alpha$ is a scaling factor, and $u(t)$ is a scalar function controlling the adaptation rate at optimization step $t$. This formulation provides two key advantages: \textbf{(1)} it eliminates inter-matrix scale conflicts by construction, ensuring stable optimization, and \textbf{(2)} it reduces the parameter count by roughly half compared to standard LoRA.

\paragraph{Initialization scheme.} To enable effective gradient flow during training, we initialize $A$ with a Kaiming distribution. 
To preserve the behavior of the pretrained model  at initialization, we require $u(0) = 0$, ensuring that the model starts from the pretrained weights. In practice, we adopt a simple ramp function for $u(t)$, namely $u(t) = \min\left(\frac{t}{T}, 1\right)$, where $T \in \mathbb{R}$ controls the adaptation rate. $u$ provides a smooth transition from the pretrained weights to the adapted model, allowing for gradual incorporation of task-specific features while maintaining stability during the early training stages.

\subsection{Efficiency Analysis of a Toy model}
\label{sec:SingLoRA_stability}
        
Similarly to Section 3, we analyze a toy example with the proposed single-matrix formulation (using only $a$ instead of both $a$ and $b$). 
Formally, $f(x) = (W_0 + u(t)aa^T)x$,
where $W_0 \in \mathbb{R}^{n \times n}$ represents the pretrained weights and $a \in \mathbb{R}^n$ is the trainable vector.
Like in Section 3, we consider a single training sample $(x,y)$ with the loss $\mathcal{L} = \frac{1}{2}\|f(x) - y\|^2$ which is minimized using a gradient descent procedure with a learning rate $\eta = \Theta(n^c)$.
The gradient $\nabla_a \mathcal{L}$ and its update at step $t$ are respectively given by, $
\nabla_a \mathcal{L} = u(t)\, [(a_{t-1}^\top x)\, (f_{t-1}(x) - y) + a_{t-1}^\top (f_{t-1}(x) - y)x] $ and by $a_t = a_{t-1} - \eta \nabla_a \mathcal{L}$.
        
\paragraph{Scaling Analysis.} For stable feature learning, we require $f_t(x) = \Theta(1)$. As $W_0$ remains frozen, this means $u(t)\, a_t\, (a_t^\top x) = \Theta(1)$. If $a_t$ has entries of order $\Theta(n^p)$, then $a_t(a_t^\top x)$ scales as $\Theta(n^{2p+1})$. For stability, this implies $2p+1 = 0$, yielding $p = -1/2$. To ensure the stability of the optimization, $a_t$ should thus maintain entries of order $\Theta(n^{-1/2})$.
        
Let us analyze the factor of the gradient update:
\begin{itemize}
    \item $\eta = \Theta(n^c)$.
    \item $\nabla_a \mathcal{L} = \Theta_n(1)$ since $(f(x) - y) = \Theta_n(1)$, as we assume the error of the model to be bounded, and $a_{t-1}^Tx = \Theta(1)$, $a_{t-1}^T (f(x) - y) = \Theta(1)$ by CLT.
\end{itemize}
The gradient update thus scales as $\Theta(n^c)$ and we require $c = -1/2$ to maintain $a_t$ with entries $\Theta(n^{-1/2})$. Therefore, setting $\eta = \Theta(n^{-1/2})$ ensures stable feature learning.
        
Unlike LoRA where balancing two matrices requires careful learning rate tuning, \slora achieves stable feature learning by design, as we can always set an appropriate learning rate scale to ensure $f_t = \Theta(1)$. 
This characteristic guarantees efficient feature learning in the infinite-width limit.

\subsection{Transformation Invariance of Low Rank Adapters}
\label{sec:SingLoRA_transformation_invariance}
Here, we extend the analysis of the previous subsection to a more general setting, and show that a model adapted with \slora and finetuned using standard optimizers, such as SGD, achieves stable feature learning.

Observing that pairs of matrices \((A_1, B_1)\) and \((A_2, B_2)\) representing the same adapter, i.e. $A_1 B_1 = A_2 B_2$, can produce different optimizer updates, Yen \textit{et al.} \cite{yen2024lora} recently introduced the notion of \textit{transformation-invariance} for low rank adapters.

\noindent \textbf{Definition 1. Transformation-Invariance}
Let \((A_1, B_1)\) and \((A_2, B_2)\) be LoRA adapters satisfying \(A_1 B_1 = A_2 B_2 \). 
An optimizer is \textit{transformation-invariant} if its updates \((\delta A_1, \delta B_1)\) and \((\delta A_2, \delta B_2)\) satisfy,
\begin{eqnarray}
    (A_1 + \delta A_1)(B_1 + \delta B_1)
   = (A_2 + \delta A_2)(B_2 + \delta B_2)
   \label{eq:transformation_invariance}
\end{eqnarray}

To illustrate how the multiplicity in LoRA's parametrization can yield different gradient descent updates, we consider two parameterizations $(A_1, B_1)$ and $(A_2, B_2)$ of a low-rank adapter related by a scaling factor $s \in \mathbb{R}$. Namely,
\begin{equation}
  A_2 = s\,A_1, \quad B_2 = \frac{1}{s}\,B_1.
\end{equation}
Defining $Z = A_1B_1 = A_2B_2$ and applying the chain rule yields $\nabla A_1 = \nabla Z B_1, \, \nabla A_2 = \nabla Z B_2$, where $\nabla P$ stands for the gradient of the loss $\mathcal{L}$ of the model with respect to the parameter $P$. We can thus rewrite,  $\nabla A_2 = \frac{1}{s} \nabla Z B_1 = \frac{1}{s} \nabla A_1$.
Therefore,
\begin{equation*}
   \delta A_1 B_1
   = - \eta \, \nabla A_1 B_1
   = - \eta \, s\,\nabla A_2 B_1 
   =  {s^2}\,\delta A_2 B_2, 
\end{equation*}
making Eq. \ref{eq:transformation_invariance} unsatisfied in the general case.
This example reveals why LoRA exhibits unstable training dynamics when using optimizers that are not transformation-invariant: when the scaling factor $s$ is much larger than 1, the matrices $A$ and $B$ (and their corresponding updates) operate at vastly different scales. 
This scale mismatch creates a fundamental problem: first-order optimizers using a single learning rate struggle to achieve stable feature learning, as they cannot simultaneously accommodate both large and small-scale updates effectively. 
This issue frequently arises during training since LoRA's matrices $A$ and $B$ are typically initialized with different scales.

\begin{table}[ht]
\centering
\resizebox{\linewidth}{!}{%
\begin{tabular}{llcccccc}
\toprule
\textbf{Dataset} & \textbf{Method} & \multicolumn{3}{c}{\textbf{RoBERTa}} & \multicolumn{3}{c}{\textbf{GPT-2}} \\
\cmidrule(lr){3-5} \cmidrule(lr){6-8}
 & & LR & Acc. (\%) & Params & LR & Acc. (\%) & Params \\
\midrule
QQP & LoRA & $2e^{-4}$ & 88.5 & 0.15M & $4e^{-4}$ & 87.9 & 1.78M \\
 & LoRA+ & $(2e^{-4},4e^{-3})$ & 89.1 & 0.15M & $(2e^{-4},4e^{-3})$ & 89.1 & 1.78M \\
 & DoRA & $5e^{-4}$ & 89.2 & 0.16M & $5e^{-4}$ & 89.2 & 1.78M \\
 & Ours & $1e^{-3}$ & 88.9 & \textbf{0.075M} & $1e^{-3}$ & 88.8 & \textbf{0.89M} \\
\midrule
QNLI & LoRA & $4e^{-4}$ & 90.9 & 0.15M & -- & -- & -- \\
 & LoRA+ & $(2e^{-4},4e^{-3})$ & 92.1 & 0.15M & -- & -- & -- \\
 & DoRA & $5e^{-4}$ & 92.1 & 0.16M & -- & -- & -- \\
 & Ours & $1e^{-3}$ & \textbf{92.2} & \textbf{0.075M} & -- & -- & -- \\
\midrule
MNLI & LoRA & $4e^{-4}$ & 85.6 & 0.15M & $2e^{-4}$ & 81.3 & 1.78M \\
 & LoRA+ & $(5e^{-5},4e^{-3})$ & 86.5 & 0.15M & $(2e^{-4},4e^{-3})$ & 82.0 & 1.78M \\
 & DoRA & $5e^{-4}$ & 86.4 & 0.16M & $5e^{-4}$ & 82.2 & 1.78M \\
 & Ours & $1e^{-3}$ & \textbf{86.5} & \textbf{0.075M} & $1e^{-3}$ & \textbf{82.5} & \textbf{0.89M} \\
\midrule
Mean & LoRA & -- & 88.3 & 0.15M & -- & 84.6 & 1.78M \\
 & LoRA+ & -- & \textbf{89.2} & 0.15M & -- & 85.6 & 1.78M \\
 & DoRA & -- & 89.2 & 0.16M & -- & 85.7 & 1.78M \\
 & Ours & -- & \textbf{89.2} & \textbf{0.075M} & -- & \textbf{85.7} & \textbf{0.89M} \\
\bottomrule
\end{tabular}%
}
\vspace{-2pt}
\caption{Accuracy of RoBERTa and GPT-2 fine-tuned on GLUE datasets with rank 8 updates. LoRA+ uses learning rates $(\mu_A, \mu_B)$. GPT-2 results on QNLI were not reported in \cite{hayou2024loraplus}.}
\label{tab: commonsense}
\vspace{-3pt}
\end{table}

\noindent \textbf{Theorem 1.}  Any transformation invariant optimizer applying the same update rule for \(A\) and \(B\) achieves efficient feature learning. \textit{A proof is presented in the appendix.}

Recent efforts \cite{yen2024lora, zhang2024riemannian} attempt to address the stability issues of LoRA by building a dedicated scale-invariant optimizer. 
In contrast, \slora formulation inherently solves those challenges and \slora  can be efficiently tuned with standard deep learning optimizers, such as SGD or Adam \cite{kingma2014adam}, without requiring special modifications or careful hyper-parameters tuning.

\noindent \textbf{Theorem 2. A gradient descent optimizer is transformation-invariant for \slora.} \textit{A proof is presented in the appendix.}

Theorem 1 and Theorem 2 guarantee the existence of a learning rate yielding stable dynamics when training \slora with first-order optimization methods. 

\subsection{Extension to Non-Square Matrices}
\label{sec:SingLoRA_non_square}
Our discussion has so far focused on square weight matrices \( W_0 \in \mathbb{R}^{n \times n} \) which are commonly used in the attention layers of transformer architecture. We now extend our approach to rectangular weight matrices \( W_0 \in \mathbb{R}^{d_{\text{in}} \times d_{\text{out}}} \). Without loss of generality, we assume \( d_{\text{in}} < d_{\text{out}} \).

Considering a low rank matrix \( A \in \mathbb{R}^{d_{\text{out}} \times r} \) we define \( A^* \in \mathbb{R}^{d_{\text{in}} \times r} \) as a truncation of \( A \) consisting of its first \( d_{\text{in}} \) rows. The adapted layer is then computed by $W_0 + A^* A^\top$.    
Training this adapter with standard optimizers preserves the stability and transformation-invariance properties demonstrated for the square case.

\noindent {\bf Theorem 3.} The generalization of \slora to non-square matrix preserves the stability and transformation-invariance properties demonstrated for the square case. \textit{A proof is presented in the appendix.}

\begin{figure}[t]
    \centering
    \includegraphics[width=0.8\linewidth]{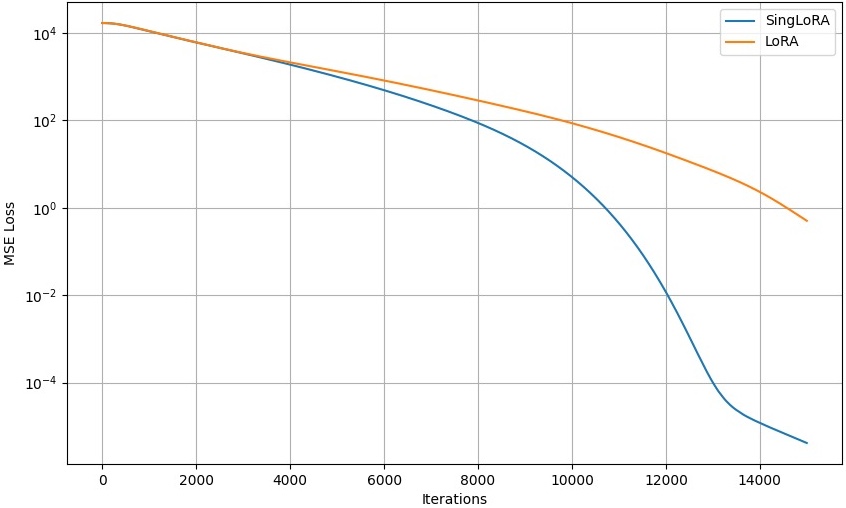}
    \vspace{-10pt}
 \caption{Synthetic experiment: convergence plot for LoRA and \slora.}
    \label{fig: conv plot}
    \vspace{-10pt}
\end{figure}

\section{On the expressiveness of \slora in Transformer Architectures}
\label{sec: Expressiveness of SingLoRA in Transformer Architectures}
In the previous section, we demonstrated that \slora improves the optimization process and yields more stable training dynamics. We now turn to investigating its expressive capacity within the Transformer architecture, which serves as the foundation of most current NLP and vision models.
We analyze how \slora's symmetric updates affect the key-query interaction in self-attention layers. 
For input $\mathbf{X} \in \mathbb{R}^{L \times d}$, the attention mechanism is computed as
    \begin{equation}
        \text{Attention}(\mathbf{Q}, \mathbf{K}, \mathbf{V}) = \text{softmax}\left(\frac{\mathbf{Q}\mathbf{K}^T}{\sqrt{d_k}}\right)\mathbf{V},  \; 
    \text{where}\; \mathbf{Q} = \mathbf{X}\mathbf{W}_q, \, \mathbf{K} = \mathbf{X}\mathbf{W}_k \,\, \text{and} \, \mathbf{V} = \mathbf{X}\mathbf{W}_v.
    \label{eq: attention}
    \end{equation}
When applying \slora, the weight matrices become,
    \begin{align}
        \mathbf{W}_q = \mathbf{W}_q^0 + \mathbf{A}_q\mathbf{A}_q^T  \; \text{and} \;
        \mathbf{W}_k = \mathbf{W}_k^0 + \mathbf{A}_k\mathbf{A}_k^T
    \end{align}
    
    Examining the key-query interaction $\mathbf{Q}\mathbf{K}^T$, we get
    \begin{align}
        \mathbf{Q}\mathbf{K}^T = \mathbf{X}[\mathbf{W}_q^0\mathbf{W}_k^{0T} + \mathbf{W}_q^0\mathbf{A}_k\mathbf{A}_k^T + \mathbf{A}_q\mathbf{A}_q^T\mathbf{W}_k^{0T} + \mathbf{A}_q\mathbf{A}_q^T\mathbf{A}_k\mathbf{A}_k^T]\mathbf{X}^T.
    \end{align}
    The crucial insight that emerges is that although \slora uses symmetric updates ($\mathbf{A}_q\mathbf{A}_q^T$ and $\mathbf{A}_k\mathbf{A}_k^T$), their interaction in the computation of attention is more general. 
    The product of two symmetric matrices $(\mathbf{A}_q\mathbf{A}_q^T)(\mathbf{A}_k\mathbf{A}_k^T)$ is not necessarily symmetric unless they commute. 
    Since there is no constraint forcing $\mathbf{A}_q\mathbf{A}_q^T$ and $\mathbf{A}_k\mathbf{A}_k^T$ to commute, \slora can learn general (non-symmetric) transformations of attention patterns despite its symmetric parameterization and does not fundamentally limit the model's ability to learn diverse attention patterns. 
    
\paragraph{Synthetic experiment.} To empirically verify this property, we implement LoRA and \slora in an attention mechanism, where they learn to approximate a target attention pattern $Z$ given input $X$. Attention scores are computed according to \ref{eq: attention}. For \slora, we define $W_q = W^0_q + A_qA_q^\top$ and $W_k = W^0_k + A_kA_k^\top$, where both $W_q, W_k \in \mathbb{R}^{128 \times 128}$. For LoRA, we define $\hat{W}_q = \hat{W}^0_q + \hat{B}_q\hat{A}_q$ and $\hat{W}_k = \hat{W}^0_k + \hat{B}_k\hat{A}_k$, also with $\hat{W}_q, \hat{W}_k \in \mathbb{R}^{128 \times 128}$. Both approaches are configured to use the exact same number of parameters for a fair comparison. 
We optimize both models using an identical AdamW configuration with a learning rate of $10^{-4}$ for $15,000$ iterations, minimizing the loss $\|XW_qW_k^TX^T - Z\|_2^2$.  As shown in Figure \ref{fig: conv plot}, \slora outperforms LoRA in both convergence speed and final approximation accuracy which drops to approximately $10^{-5}$ while LoRA remains around $10^{-2}$, in fewer iterations. We verified these results with 1K different random seeds to sample $Z$ and $X$. This experiment shows how \slora maintains expressiveness in attention mechanisms while enhancing optimization dynamics and performance.

\section{Experiments}
We conduct extensive experiments to evaluate \slora in low-rank adaptation for linguistic and visual tasks.
Additional experiments and ablative studies on the ramp-up function, and influence of ranks are presented in the appendix. Code will be released upon publication.

\subsection{Language Models}
To evaluate \slora in Natural Language Processing, we consider selected comprehension and reasoning tasks from the \textbf{G}eneral \textbf{L}anguage \textbf{U}nderstanding \textbf{E}valuation (GLUE) benchmark\cite{wang2018glue}. We strictly follow the NLP's experimental protocol of LoRA+, which similarly to \slora addresses LoRA’s training stability issues and thus serves as a key benchmark. To fairly quantify algorithmic differences rather than hyper-parameter tuning advantages, we adopt their training/evaluation codebase, model architectures, modified layers, optimization settings, and training duration. We also compare to DoRA \cite{liu2024dora}, a recent state-of-the-art variation of LoRA.

\paragraph{RoBERTa-base and GPT-2.} 
We first evaluate each approach based on its ability to fine-tune smaller language models - RoBERTa-base and GPT-2 - on the MNLI, QQP, and QNLI tasks from the GLUE benchmark.
Following the LoRA+ setup, we set $r = \alpha=8$.
We use \(u(t)=\min(\frac{t}{10^3}, 1) \) where $t$ is the training step.
Table~\ref{tab: commonsense} summarizes the accuracy in these tasks. Accuracies reported for LoRA and LoRA+ are directly taken from the original paper of LoRA+ \cite{hayou2024loraplus}. Results show that \slora outperforms LoRA, achieving a $0.9\%$ mean accuracy improvement for RoBERTa and $1.1\%$ for GPT-2.
It also achieves slightly better performance than LoRA+ and DoRA, while using only half the number of trainable parameters of both baselines.
Note that while LoRA and LoRA+ explore multiple learning rates (5 for LoRA and 25 for LoRA+) and report results using carefully selected learning rates for each dataset and model, \slora employs a single learning rate across all experiments.
This indicates that our method's stability reduces the need for extensive hyperparameter tuning, including the exhaustive grid search required for LoRA+.
We further analyze the robustness to learning rate variations in Subsection~\ref{subsec:exp_stability}.

\paragraph{Llama 7B.}
\begin{table}[t]
    \centering
    \begin{tabular}{lcccc}
        \toprule
        \textbf{Method} & \textbf{LoRA} & \textbf{LoRA+} &  \textbf{DoRA} & \textbf{\slora} \\
        \midrule
        Accuracy  & 89.1 & 90.2 & 90.6 & \textbf{91.3}\\
        \# Params  & 20M & 20M & 21M & 12M \\        
        \bottomrule
    \end{tabular}
\vspace{2pt}
\caption{Accuracy of Llama tuned on MNLI with \slora and baselines with ranks 8.}
\vspace{-3pt}
\label{tab: llama}
\end{table}

To further validate our approach, we fine-tune a large language model (LLM), LLaMA-7B, on the MNLI task.
Table \ref{tab: llama} shows that \slora outperforms LoRA (by more than $2\%$), LoRA + (by more than $1\%$) and DoRA, while reducing the number of training parameters by $40\%$.
Since fine-tuning LLMs such as LLaMA is one of the most common applications of low-rank adapters, this result underlines the practical advantages of our approach.

\subsection{Image Generation}
\label{sec:image_diffusion}
To showcase the versatility of \slora in architectures and modalities, we evaluated its effectiveness in diffusion-based image generation, where LoRA is commonly used to fine-tune models in small-scale datasets of specific subjects.
We consider Stable Diffusion V1.5 \cite{rombach2022high}, a popular image diffusion model.

\begin{figure}[t]
    \centering
    \includegraphics[width=0.6\linewidth]{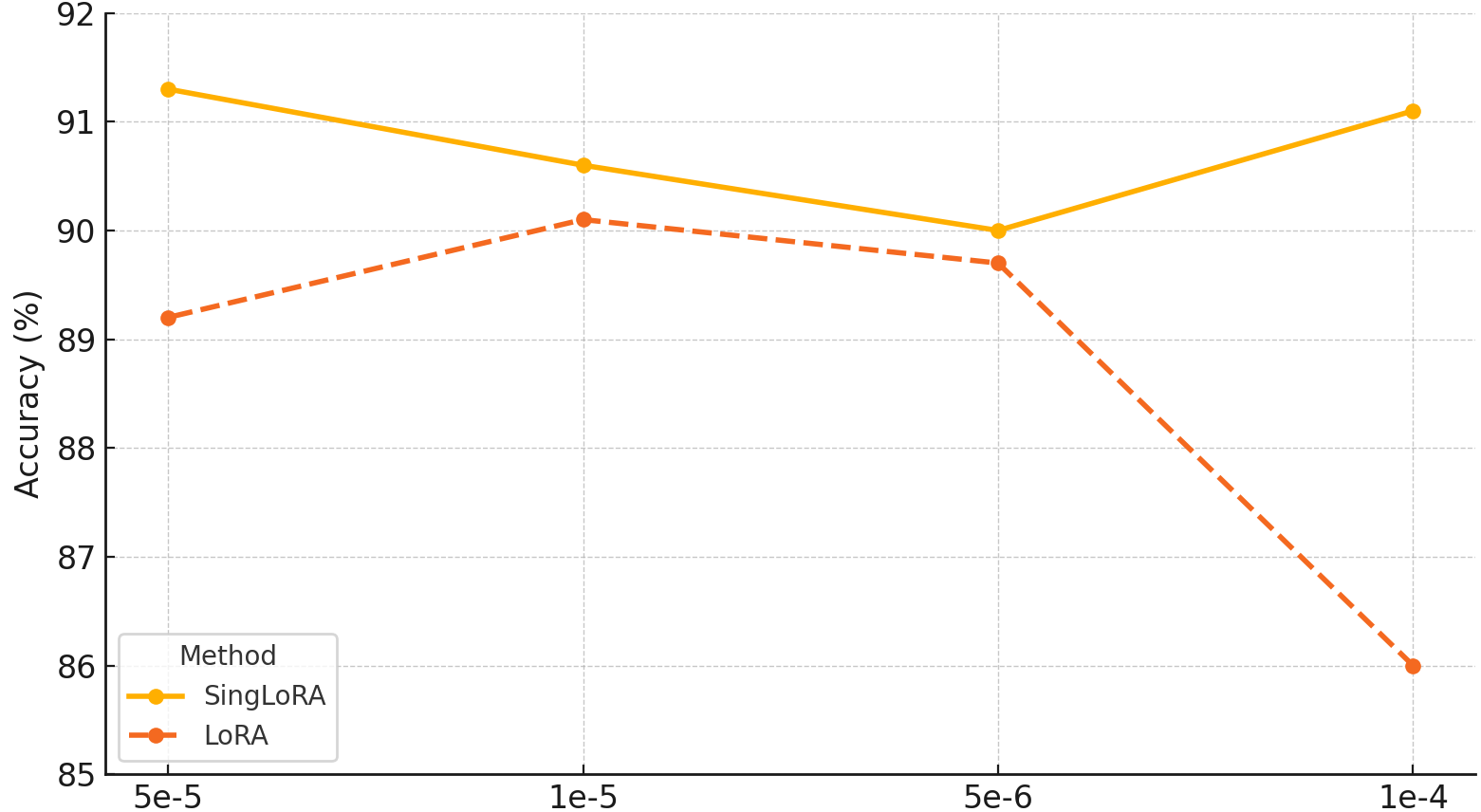}
    \vspace{-7pt}
 \caption{Accuracy of Llama-7B fine-tuned on MNLI across different learning rates. 
 The plot compares the stability of LoRA and \slora under varying learning rates, demonstrating that \slora’s accuracy fluctuates by approximately 1\%, while LoRA’s performance varies by 4.8\%. 
 These results highlight the robustness of \slora's optimization dynamics.}
    \label{fig:stability}
    \vspace{-10pt}
\end{figure}

\begin{figure*}[t]
    \centering
    \includegraphics[width=0.9\linewidth]{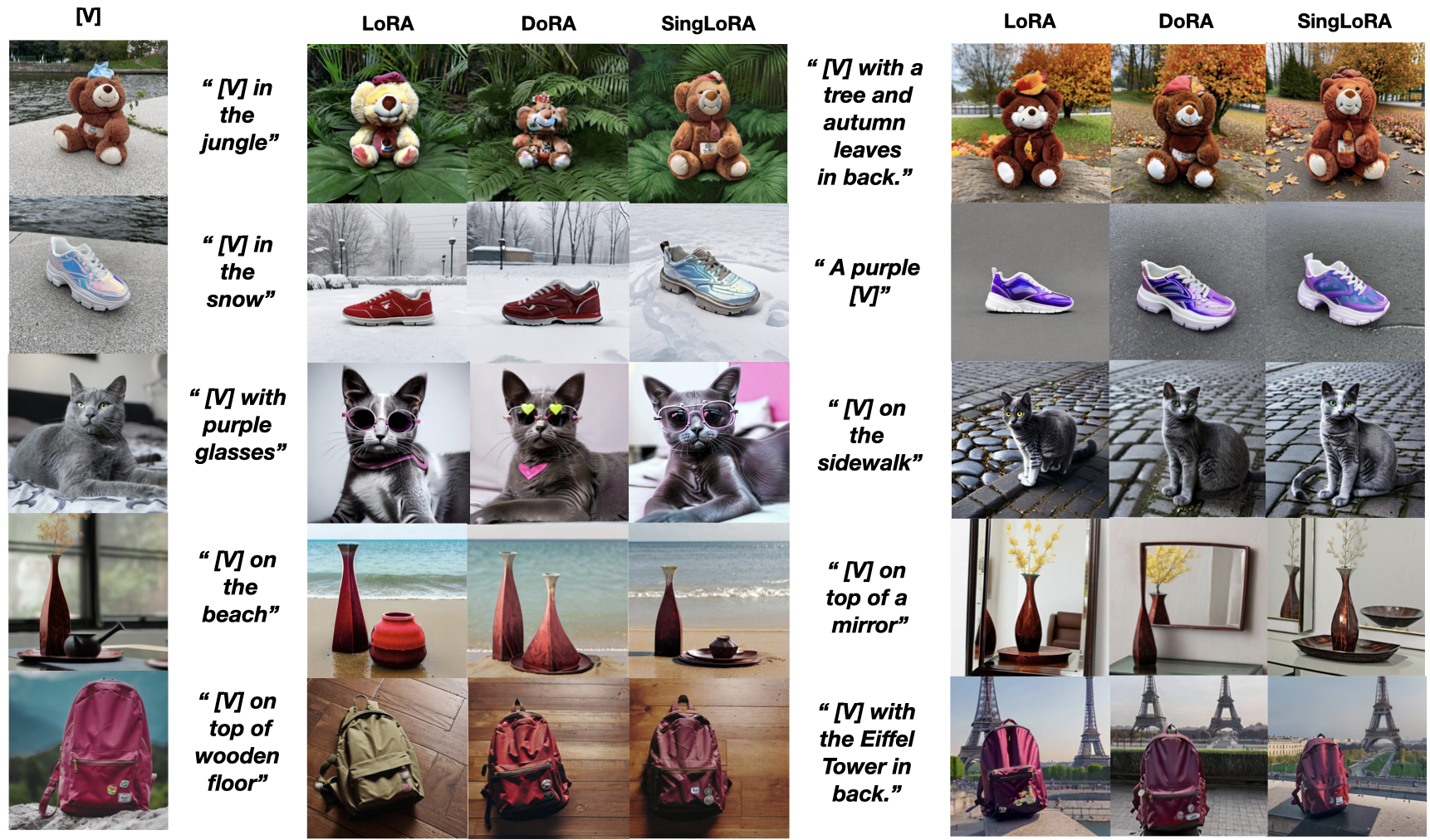}
    \vspace{-10pt} 
    \caption{Qualitative comparison of LoRA, DoRA and \slora on Dreambooth.}
    \vspace{-10pt}
    \label{fig: dreambooth}
\end{figure*}

\begin{table*}[t]
    \centering
    \begin{tabular}{lccccc}
        \toprule
        \textbf{Method} & \textbf{CLIP Image} & \textbf{CLIP Text} & \textbf{DINO Similarity} & \textbf{Rank} & \textbf{\# Params} \\
        \midrule
        LoRA    & 0.677 & 0.319 & 0.143 & 8  & 0.9M \\
        LoRA+    & 0.688 & 0.315 & 0.150 & 8  & 0.9M \\
        DoRA    & 0.687 & 0.317 & 0.148 & 8  & 0.9M \\
        \slora & 0.677 & 0.318 & 0.141 & 8 & 0.45M \\
        \slora & 0.690 & 0.317 & 0.151 & 16 & 0.9M \\
        \bottomrule
    \end{tabular}
    \vspace{-2pt}
\caption{Performance of Stable diffusion V5 finetuned in Dreambooth with same number of parameters. 
Similarity of the generated image with the originals was measured using CLIP Image and DINO Similarity. 
Prompt fidelity is evaluated with CLIP Text.}
    \label{tab:diffusion_results}
    \vspace{-3pt}
\end{table*}

\paragraph{Dreambooth.} We benchmark the approaches on DreamBooth \cite{ruiz2023dreambooth}, a known dataset with 30 classes of objects and animals, each containing 4–5 training images and 25 evaluation prompts.
Following standard practice, we train each model for 400 iterations using the template prompt "\texttt{a photo of a sks <class>}", where \texttt{"sks"} is a rare English token. 
This setup allows the model to learn new object representations while retaining its general capacities.
We finetune the query and key projections of the attention matrices in the U-Net component of Stable Diffusion.
For all methods, training is conducted for 400 epochs using a learning rate of $10^{-3}$.
Figure \ref{fig: dreambooth} presents a qualitative comparison between the methods, illustrating how our adaptation approach enhances subject learning. 
For instance, note that the shoe in the second row retains its iridescent color, whereas other methods fail to do so.
Quantitatively, Table~\ref{tab:diffusion_results} demonstrates that \slora achieves a $5.4\%$ improvement in the DINO \cite{caron2021emerging} similarity score compared to DoRA, indicating better object resemblance, while maintaining prompt fidelity measured by the CLIP \cite{radford2021learning} Text score.
Additional image generation experiments are presented in the appendix.

\subsection{Stability of \slora}\label{subsec:exp_stability}
To validate the optimization stability analysis of \slora presented in Section~\ref{sec:SingLoRA_transformation_invariance}, we compare its performance against LoRA in a range of learning rates.
Specifically, we fine-tune Llama-7B in MNLI using learning rates ranging from $5\cdot10^{-5} \text{ to } 10^{-4}$.
As shown in \Cref{fig:stability}, the precision of \slora fluctuates by approximately $1\%$, whereas LoRA exhibits a larger variation of up to $4.8\%$.
These empirical results validate our theoretical findings, demonstrating that the design of \slora inherently improves learning stability.
In practice, this stability translates to simpler hyperparameter tuning, as our method maintains strong performance without requiring extensive searches for an optimal learning rate.
Together with previous experiments, these findings highlight that \slora not only offers a more efficient and accurate parameterization, but also ensures robust convergence in practical settings.

\section{Conclusion}
We introduced \slora, a novel formulation of low-rank adaptation that learns and employs a single matrix instead of two.
Through theoretical analysis, we demonstrated that the proposed design inherently eliminates the inter-matrix scale disparities present in LoRA and guarantees stable feature learning without requiring special optimizers or careful hyperparameter tuning.
Extensive experiments on language and vision tasks validated these benefits, consistently demonstrating improved performance with fewer trainable parameters than LoRA and its variants.
Since our approach is complementary to various LoRA's variants, suchs as DoRA \cite{liu2024dora}, a promising direction is to explore their integration, harnessing their independent strengths to further enhance efficiency and performance.

\newpage
{
    \bibliographystyle{plain}
    \bibliography{bib}
}
\newpage
\appendix
\input{appendix}
\newpage

\end{document}

%% file: appendix.tex
\section*{Supplementary}
\section{Additional experiments}
\subsection{Initialization and choice of T} 
\begin{figure}[h]
    \centering
    \includegraphics[width=0.8\linewidth]{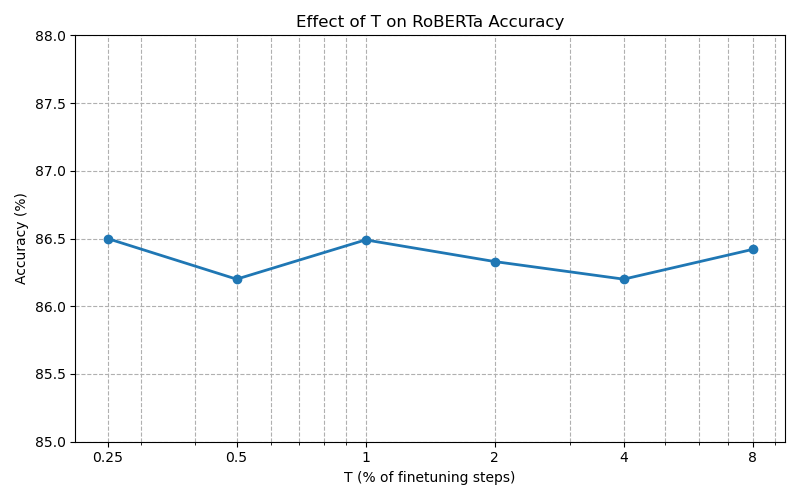}
 \caption{Ablation study on the choice of the hyperparameter $T$. The experiments shows that \slora is robust to a wide range of $T$ which thus does not require extensive hyperparameter search.}
    \label{fig: ablation}
\end{figure}

In LoRA, $A$ is initialized using a Kaiming distribution while $B$ is set to zero. This initialization scheme ensures that the model is equivalent to the pretrained one at the beginning of finetuning (since $AB=0$), while maintaining effective gradient flow. For SingLoRA, we adopt a similar approach by initializing $A$ with a Kaiming distribution. To achieve equivalence to the original model at the beginning of finetuning, we rely on a simple parametric ramp-up function defined as $t/T$, where $t$ represents the current training step and $T$ is a hyperparameter, and consider the adapter,
\begin{equation}
    u(t) AA^\top
\end{equation}
To analyze the robustness of this initialization scheme, we conduct an ablation study focusing on the ramp-up function and the hyperparameter $T$. Figure \ref{fig: ablation} demonstrates that \slora is robust to the choice of $T$, showing comparable of RoBERTa's performance in GLUE across a wide range of $T$ values (from $0.5\%$ to $8\%$ of the total number of training steps). In our main experiments, we set $T$ equal to $1\%$ of the total number of training steps.

\subsection{Additional visual experiment: human faces}

\begin{figure}[h]
    \centering
    \includegraphics[width=0.8\linewidth]{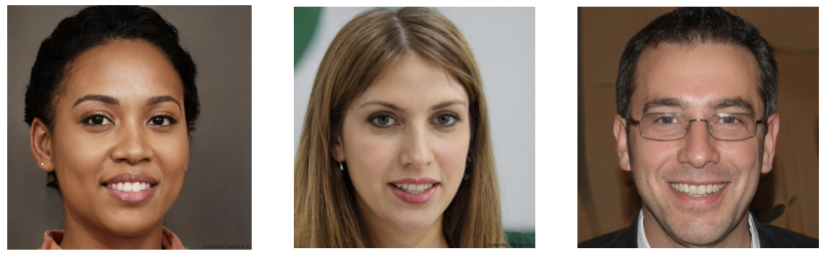}
 \caption{Samples from the dataset used in our experiment. The dataset was automatically generated with a state-of-the-art image generation model trained on faces available in \nolinkurl{https://this-person-does-not-exist.com/en}.}
    \label{fig: faces}
\end{figure}

To further analyze the ability of \slora in visual tasks, we follow [20] and benchmark the adapters on a dataset that includes 40 human faces. This experiment quantifies the expressive power of adapters and their ability to learn complex details present in human faces. Figure \ref{fig: faces} shows a subset of the dataset.
Following standard practice, we train each adapter for 1500 iterations using the template prompt ``\texttt{a photo of a sks human}'', where \texttt{sks} is a rare English token. Each adapter has the same number of trainable parameters.
We finetune the query and key projections of the attention matrices in the U-Net component of Stable Diffusion.
Table~\ref{tab: faces} shows an improvement in the DINO similarity score compared to DoRA, indicating a better similarity to the reference image.

\subsection{Hardware}
All experiments were performed on a single NVIDIA A40 GPU with a memory of 48GB.

\begin{table}[h]
    \centering
    \begin{tabular}{lcccc}
        \toprule
        \textbf{Method} & \textbf{LoRA} &  \textbf{DoRA} & \textbf{\slora} \\
        \midrule
        DINO Similarity  & 0.463 & 0.471 & 0.501\\
        Rank  & 8 & 8 & 16 \\        
        \bottomrule
    \end{tabular}
\caption{DINO Similarity of Stable Diffusion v1.5 tuned on a specific face with the reference face. \slora and baselines uses the same number of trainable parameters.}
\label{tab: faces}
\end{table}

\newpage
\section{Proofs}
\label{sec:appendix-proofs}
In this section we demonstrate the theorem introduced in Sections 4 and 5.
We rely on the following sufficient conditions for transformation invariance (see [20]),
\begin{eqnarray}
    \textbf{(i)}  \quad \quad \, \quad \delta A_1 B_1 &=& \delta A_2 B_2 \cr
     \textbf{(ii)} \quad \quad \quad A_1 \delta B_1 &=& A_2 \delta B_2  \cr
    \textbf{(iii)} \quad \quad \, \, \delta A_1 \delta B_1 &=& \delta A_2 \delta B_2  
\label{eq: sufficient condition} 
\end{eqnarray}

\subsection{Theorem 1}
\noindent Any transformation invariant optimizer applying the same update rule for \(A\) and \(B\) achieves efficient feature learning.

\textit{Here for completeness we present the proof proposed in [20]}.
\textbf{Proof:}

Let $\|\mathbf{A}_1\| = \Theta(n^a)$, $\|\mathbf{B}_1\| = \Theta(n^b)$, $\|\nabla \mathbf{Z}\| = \Theta(n^c)$, and $\eta = \Theta(n^d)$, where $\eta$ is the learning rate and $n$ denotes the network width. Since $\mathbf{Z} = \mathbf{A}_1 \mathbf{B}_1^\top$, by the chain rule, we have $\nabla \mathbf{Z} = \nabla \mathbf{Z}^\top = \nabla \mathbf{Z} \mathbf{B}_1 \mathbf{A}_1^\top$. Given the symmetry of the update rule, the updates satisfy:
\[
\|\delta \mathbf{A}_1\| = \Theta(n^{x+a+(y+1)b+c+d}), \quad \|\delta \mathbf{B}_1\| = \Theta(n^{x+b+(y+1)a+c+d}).
\]

Assuming the update rule is invariant under scalar reparameterization, we compare two equivalent decompositions $\mathbf{A}_2 = n^s \mathbf{A}_1$ and $\mathbf{B}_2 = n^{-s} \mathbf{B}_1$, giving:
\[
\|\delta \mathbf{A}_1\| \|\mathbf{B}_1\| = \|\delta \mathbf{A}_2\| \|\mathbf{B}_2\|.
\]

From this, it follows:
\[
x a + (y+1)b + z c + d = x(a + s) + (y+1)(b - s) + z c + d,
\]
which simplifies to:
\[
x s - (y+1)s = 0 \quad \forall s \Rightarrow x = -1.
\]

Hence, we deduce:
\[
\|\delta \mathbf{A}_1\| \|\mathbf{B}_1\| = \Theta(n^{a + (y+1)b + c + d}) = \Theta(n^{a + b + yc + d}).
\]

Similarly, we find:
\[
\|\mathbf{A}_1\| \|\delta \mathbf{B}_1\| = \Theta(n^{a + (y+1)b + c + d}) = \Theta(n^{a + b + yc + d}).
\]

Given that these expressions are equal, the update process enables efficient feature learning:
\[
\|\delta \mathbf{A}_1\| \|\mathbf{B}_1\| = \|\mathbf{A}_1\| \|\delta \mathbf{B}_1\| = \Theta(1),
\]
by selecting a proper learning rate $\eta = \Theta(n^d)$, where $x = -1$ is fixed and $d$ is chosen accordingly.

\subsection{Theorem 2}

\noindent A gradient descent optimizer is transformation-invariant for \slora.

\textbf{Proof:}

We prove that the three sufficient conditions for transformation invariance hold.
\textbf{Proof of (i):} Assume two parametrizations of a LoRA adapter, $A_1$, $A_2 \in R^{n \times r}$ with identical ranks such that $A_1A_1^\top = A_2A_2^\top$. 
From the Polar Decomposition Theorem, there exists an orthogonal matrix $Q \in \mathbb{R}^{r \times r}$ such that $A_1 = A_2 Q$.  
Therefore, defining $X=A_1 A_1^\top$ and using the chain rule lead to,
\begin{align}
    \delta A_1 A_1^\top &= -\eta \nabla A_1 A_1^\top = -2\eta \nabla{Z} A_1 A_1^\top  = -2\eta \nabla Z A_2 Q Q^\top A_2^\top \cr
    &= -2\eta \nabla Z A_2 A_2^\top= -\eta \nabla A_2 A_2^\top = \delta A_2 A_2^\top.
\end{align}
The first sufficient condition for transformation invariance \ref{eq: sufficient condition} is therefore satisfied. 

Condition \textbf{(ii)} holds by symmetry with the condition \textbf{(i)}.

\textbf{Proof of (iii): $\delta A_1\,\delta A_1 = \delta A_2\,\delta A_2$}
For the gradients with respect to $A$, the chain rule gives


\begin{equation}
    \nabla_{A_1} \mathcal{L} = \nabla_Z \mathcal{L} \cdot A_1 \quad \nabla_{A_2} \mathcal{L} = \nabla_Z \mathcal{L} \cdot A_2
    \label{eq:gradients}
\end{equation}

where \(Z = A_i A_i^\top\) is the output of the adapted layer. Using the former relation and the Polar Theorem, there exists a matrix $Q \in \mathbb{R^{r\times r}}$ such that,
\begin{align}
    \nabla_{A_1}\mathcal{L} &= \nabla_Z \mathcal{L}\,A_1 = (\nabla_Z \mathcal{L} \,A_2)Q = \nabla_{A_2} \mathcal{L} \, Q
\end{align}

Hence, the update for $A$ becomes
\begin{equation}
    \delta A_1 = -\eta\,\nabla_{A_1} \mathcal{L} = -\eta\,\nabla_{A_2} \mathcal{L} \,Q = \delta A_2\,Q.
\end{equation}

Now, consider the product of the updates:
\begin{align}
    \delta A_1\,\delta A_1^\top &= (\delta A_2\,Q)(\delta A_2\,Q)^\top = \delta A_2\,(Q Q^\top)\,\delta A_2 ^\top \cr &= \delta A_2\,\delta A_2^\top,
\end{align}
since \(Q Q^\top = I\). This completes the proof of \textbf{(ii)}.

\subsection{Theorem 3}

\noindent The generalization of \slora to non-square matrix preserves the stability and transformation-invariance properties demonstrated for the square case.

\textbf{Proof:}

Recall that for a rectangular weight matrix, 
\[
W_0 \in \mathbb{R}^{d_{\text{in}} \times d_{\text{out}}} \quad (d_{\text{in}} < d_{\text{out}}),
\]
a low-rank adapter is defined via a matrix $A \in \mathbb{R}^{d_{\text{out}} \times r}$,
with its truncation $A^* \in \mathbb{R}^{d_{\text{in}} \times r}$ formed by the first \(d_{\text{in}}\) rows of \(A\). 

Suppose that there exist two matrices
$A_1, A_2 \in \mathbb{R}^{d_{\text{out}} \times r}$ such that their truncations satisfy
\[
A_1^* A_1^\top = A_2^* A_2^\top.
\]
For clarity, divide each \(A_i\) as
\[
A_i = \begin{bmatrix} X_i \\ Y_i \end{bmatrix},\quad i=1,2,
\]
where $X_i \in \mathbb{R}^{d_{\text{in}} \times r}$ and $Y_i \in \mathbb{R}^{(d_{\text{out}}-d_{\text{in}}) \times r}$.



By definition, the equality $A_1^* A_1^\top = A_2^* A_2^\top$ implies,

\begin{equation}
\begin{aligned}
    X_1 X_1^\top &= X_2 X_2^\top \\
    X_1 Y_1^\top &= X_2 Y_2^\top.
\end{aligned}
\end{equation}




Using the polar theorem, $X_2$ and $Y_2$ admit a polar decomposition:
\begin{equation}
\begin{aligned}
    X_2 &= X_1 Q \\
    Y_2 &= Y_1 Q.
\end{aligned}
\label{eq:polar_XY}
\end{equation}
Where $Q$ is a symmetric orthogonal matrix.

Denote the gradient descent updates for \(A_i\) as,
\[
\delta A_i = \begin{bmatrix} \delta X_i \\ \delta Y_i \end{bmatrix},\quad i=1,2.
\]

By equation \eqref{eq:polar_XY} we have,

\begin{equation}
    \delta A_2 = \begin{bmatrix} \delta X_1 \, Q \\ \delta Y_1 \, Q \end{bmatrix}.
    \label{eq:delta_A2}
\end{equation}

We now demonstrate that the three sufficient conditions for transformation invariance hold.
\textbf{Proof of (i) and (ii): \(A_1\, \delta A_1^\top = A_2\, \delta A_2^\top\).}
Using equations \eqref{eq:polar_XY} and \eqref{eq:delta_A2}, we get
\begin{align}
    A_2^{*}\, \delta A_2^\top &= X_1 Q \begin{bmatrix} \delta X_1 Q \\ \delta Y_1 Q \end{bmatrix}^\top \cr &= [X_1 Q Q^\top \delta X_1^\top | X_1 Q Q^\top \delta Y_1^\top ] \cr&= [X_1 \delta X_1^\top | X_1  \delta Y_1^\top ] = X_1  \begin{bmatrix} \delta X_1 \\ \delta Y_1 \end{bmatrix}^\top \cr
    &= A_1^{*} \delta A_1^\top
\end{align}
The proof for \( A_1^{*\top} \delta A_1 = A_2^{*\top} \, \delta A_2 \) is symmetric.

\textbf{Proof of (iii): \(\delta A_1\, \delta A_1^\top = \delta A_2 \, \delta A_2^\top\).}
Using equations \eqref{eq:polar_XY} and \eqref{eq:delta_A2}, we get
\begin{align}
    \delta A_2^{*}\, \delta A_2^\top &= \delta X_1 Q \begin{bmatrix} \delta X_1 Q \\ \delta Y_1 Q \end{bmatrix}^\top \cr &= [\delta X_1 Q Q^\top \delta X_1^\top | \delta X_1 Q Q^\top \delta Y_1^\top ] \cr&= [\delta X_1 \delta X_1^\top | \delta X_1 \delta Y_1^\top ]  = \cr
    &= \delta A_1^{*} \delta A_1^\top
\end{align}
This completes the proof of \textbf{iii}.